# A Physics-based Domain Adaptation Framework for Modelling and Forecasting Building Energy Systems

Zack Xuereb Conti[1,2], Ruchi Choudhary[1,2], Luca Magri[1,3]

[1]Data-Centric Engineering program, The Alan Turing Institute, 96 Euston Road, London NW1 2DB, United Kingdom

[2]Department of Engineering, University of Cambridge, Trumpington Street, Cambridge CB21PZ, United Kingdom

[3]Department of Engineering, Imperial College London, South Kensington Campus, London SW7 2AZ, United Kingdom

Abstract

State-of-the-art machine-learning based models are a popular choice for modelling and forecasting energy behaviour in buildings because given enough data, they are good at finding spatiotemporal patterns and structures even in scenarios where the complexity prohibits analytical descriptions. However, their architecture typically does not hold physical correspondence to mechanistic structures linked with governing physical phenomena. As a result, their ability to successfully generalize for unobserved timesteps depends on the representativeness of the dynamics underlying the observed system in the data, which is difficult to guarantee in real-world engineering problems such as control and energy management in digital twins. In response, we present a framework that combines lumped parameter models (LPM) in the form of linear time-invariant (LTI) state space models (SSM) with unsupervised reduced order modeling (ROM) in a subspace-based domain adaptation (SDA) approach, which is a type of transfer learning (TL) technique. Traditionally, SDA is adopted for exploiting labelled data from one domain to predict in a different but related target domain for which labelled data is limited. We introduce a novel SDA approach where instead of labelled data, we leverage the geometric structure of the LTI SSM governed by well-known heat transfer ordinary differential equations (ODE) to forecast for unobserved timesteps beyond available measurement data by geometrically aligning the physics-derived and data-derived embedded subspaces closer together. In this initial exploration we evaluate the Physics-based SDA framework on a demonstrative heat conduction scenario by varying the thermophysical properties of the source and target systems to demonstrate the transferability of mechanistic models from Physics to observed measurement data.

Impact Statement

This paper addresses generalisation limitations with traditional data-driven modelling methods applied to building energy forecasting by introducing a transfer learning approach that combines Physics-based lumped parameter models in the form of linear state space models and unsupervised reduced order modelling methods. Instead of learning black-box models whose generalizability to forecast for unobserved timesteps depends wholly on the representativeness of underlying dynamics in the data, our aim is to leverage the governing structure of low-rank state space models in a domain adaptation framework. State space models are well-established for building energy forecasting and control purposes and are straightforward to derive from well-known energy transfer ODEs.

# 1. Introduction

This paper addresses generalisation limitations of traditional data-driven methods when applied to building energy modelling and forecasting. We present a novel transfer learning (TL) framework that combines Physics-based models widely used in the energy community for model predictive control, with Domain Adaptation (DA), which is a TL technique typically adopted to leverage pre-trained models for prediction across different but related tasks.

## 1.1. Building Energy Modelling and Forecasting

The urgency of an ongoing climate crisis in addition to a rising demand for thermal human comfort highlights the need to mitigate energy demand in buildings. Over the last decade, building energy control and operation strategies coupled with innovations in smart grid infrastructures at the urban scale, and controllable energy systems such as heating, ventilation and air-conditioning (HVAC), battery storage and renewable energy generation at localised building scale, has emerged as a significant approach towards manipulating the energy demand in daily post occupancy operations. On the other hand, advanced energy analysis and optimization methods in preoccupancy stages assist the mitigation of critical design factors influencing both the passive and operative energy demand generation. More recently, Digital Twins that couple energy models with measurement data provide a framework to optimize energy systems of buildings in real-time.

Building energy modelling and forecasting plays a vital role in building energy mitigation during both pre and post occupancy stages. A significant volume of past and recent literature (Bourdeau et al., 2019; Li & Wen, 2014a; Tardioli et al., 2015) categorize techniques for building energy modelling and forecasting into three main approaches: *physics-based*, *data-driven* and *hybrid* models. Physics-based models, also known as white-box models (Tardioli et al., 2015), are derived directly from mechanistic knowledge of Physics principles. Data-driven models, also known as black-box models (ASHRAE, 2009; Tardioli et al., 2015), rely on timeseries data and machine learning (ML) algorithms to accurately learn input-output relationships; these include classic linear regression and more advanced data-driven methods such as random forest (RF) (Z. Wang et al., 2018), support vector machine regression (SVR) (Shao et al., 2020) and long short-term memory (LSTM) (Sehovac et al., 2019; Somu et al., 2020; W. Wang et al., 2019). Hybrid models, also referred to as grey-box models (Tardioli et al., 2015), are a combination of both physics-based and data-driven models that use simplified physical descriptions of building energy systems via low-rank models and subsequently use data to infer the model parameters via system identification methods; these include Resistance-Capacitance (RC) thermal networks and state space models (SSM) (Amara et al., 2015; Candanedo et al., 2013; Fateh et al., 2019).

While Physics-based models are fully interpretable, they often become infeasible for complex energy systems such as largescale building with several thermal zones because they require large numbers of parameters to be specified, coupled with a computational expense to solve. In such cases, it is typical to either simplify the energy system via hybrid modelling (Goyal & Barooah, n.d.) or learn an input-output black-box surrogate model via data-driven modelling (Rätz et al., 2019). The latter has become a widely popular approach for building energy modelling and forecasting due to the recent emergence of ML techniques where given enough data, they are good at finding spatiotemporal patterns and structures even in scenarios where the complexity prohibits analytical descriptions (Willard et al., 2020). In the building energy community, vast literature has demonstrated tha tML-based energy modelling methods outperform classic statistical methods in terms of accurately capturing spatiotemporal structures from data for short term forecasting (Sehovac et al., 2019; Shao et al., 2020; Somu et al., 2020; W. Wang et al., 2019; Z. Wang et al., 2018). However, ML-based models are prone to caveats that often impede their application to real-world engineering applications such as real-time monitoring in digital twins (Karpatne et al., 2017). In this paper we identify these caveats as i) model-interpretability, ii) model-generalisability, iii) and data-dependency.

The underlying structure of ML-based models such as for example deep neural networks, typically take the form of a so-called multi-layer perceptron; a convoluted graph network connecting inputs to outputs. The graph does not hold physical correspondence to mechanistic structures linked with governing phenomena thus render such models difficult to interpret. The network topology and corresponding hyperparameters are typically determined from data using black-box identification techniques. Consequently, the ability for the black-box model to generalise for unseen inputs is heavily influenced by how well the governing behaviour of the true system underlying the data is represented in the dataset used for training (not simply on the size of the dataset, as is commonly assumed). If for example, certain occurrences in the dataset are observed more frequently than is characteristic of the true system's dynamical behaviour then, training on such data introduces bias to the model-learning, resulting in poor predictive generalization (Kouw & Loog, 2018). Since there is no easy way to determine how complete a governing behaviour is represented in a data set or how to control the 'representativeness'; black-box models rely on the size of the dataset to secure generalisation robustness, in a hit a mis approach. Thus, we can argue that improving the generalisability of typical black-box models for energy forecasting may result in a data-intensive task, which may not be suitable for applied scenarios where for example, obtaining measurement data is challenging or not feasible.

While all caveats need to be addressed, in this paper we focus on addressing *generalisability*. Specifically, we aim to improve the generalisation of building energy models such that we can forecast for unobserved timesteps given building measurement timeseries data. We present an approach that combines Physics-derived lumped parameter models (LPM) widely adopted in the building energy community for simplified thermal modelling, with DA which is a type of TL technique to leverage pre-trained models for prediction across different but related tasks. In summary, our goal is to forecast for unobserved measurement data by leveraging the generalizability innate in the governing dynamics derived from well-known ordinary differential equations (ODE) of heat transfer, mechanistically represented as LPMs, even if the latter describes the observed system approximately.

### 1.2. Lumped Parameter Models for building thermal modelling

Thermal modelling of real-world building scenarios may involve high-order intricacies. In scenarios where real-time computation is of essence, such as model predictive control (MPC) in Digital Twins, or iterative computer-based optimisation, it becomes infeasible to model and solve such models analytically nor numerically. Instead, LPMs are a popular alternative because they balance out computation time and accuracy while retain mechanistic integrity of the Physics to a reasonable degree. Such model simplification methods are also referred to as low-rank models. A popular LPM class of thermal modelling in buildings are RC networks that use an electric circuit analogy to represent the principal energy flow and energy transfer phenomena governing the energy behaviour at varying scales; across thermal zones and building components whose behaviour is influenced by thermal dynamics (Fayazbakhsh et al., 2015; Koeln et al., 2018; Lin et al., n.d.), and across urban scales (Bueno et al., 2012). An implementation example of an RC network can be found in the demonstrative example in section 2. When representing thermal dynamics of a building as an RC network, the capacitors represent thermal capacitance of the air within a zone or the material of a building component, while the resistors represent thermal resistance of the medium between adjacent thermal capacitances (Amara et al ., 2015; Goyal & Barooah, 2012). Once assembled, the network of resistors and capacitors reflect an adjacency network of thermal zones and components via interconnecting energy flow paths. The heat transfer dynamics at each capacitance node can be described by ordinary differential equations, whose parameters are either derived from well-known laws of Physics or inferred from data through system identification techniques.

In building control applications such as MPC, RC networks are typically formulated in a more mathematically standardised formulation where the coupled system of ODEs in the RC network are written as a linear time-invariant state space model (LTI SSM), which is a more compact format for representing the mechanistic structure of the transient thermal dynamics that for example relates the control signals to the space temperatures and humidity of each zone. Various applications of LTI SSMs to building energy modelling have shown that linear models sufficiently balance out prediction accuracy and model simplicity, as required for MPC (Candanedo et al., 2013; Li & Wen, 2014b; Picard et al., 2015). Further literature such as Goyal & Barooah, 2012b) present a reduced order modelling technique catered for reducing the order of energy SSMs with large number of states, while maintaining a reasonable prediction accuracy. In the case of non-linearities present in the building energy model due to for example longwave radiation exchange, absorption of incident solar radiation at the facade, or convective heat transfer, Picard, D., Jorissen, F., & Helsen, L., (2015) suggest linearisation techniques by retaining the non-linear terms in their state space formulation.

Despite being linear and approximate, we adopt LTI SSMs in our framework because of their great convenience for control design and estimation, which is justified further by their popularity in the building energy community and by their reasonable robustness in preserving the important dynamics, as illustrated in the above literature. Unlike purely data-driven black-box models, LTI SSMs facilitate access to the mechanistic structure responsible for governing the observed dynamics, even when physics only describes the dynamics of the system at hand, approximately. It is the innate generalisability thanks to their mechanistic structure, which we aim to leverage using TL.

### 1.3. Transfer Learning

Transfer learning breaks the notion of traditional ML where the training and testing data must come from the same feature space and similar distributions. In fact, in scenarios where training data is limited, TL strategies can be useful to adapt a trained model from a different but related task. TL methods can be categorized depending on the availability of labelled data in the source and target domains, as follows: a) inductive TL, where both source and target labelled data are available; b) transductive TL, where labelled data is only available at the source domain but limited to no labelled data is available at the target domain; and c) unsupervised TL is similar to inductive learning but assumes no labelled data available. A full description of each of these categories, and further subcategories can be found in an extended survey on TL techniques (Pan & Yang, 2010).

In the building energy community, the application of TL is recent. Most applications aim to overcome the dependency on historical data by transferring or adapting data-driven models trained for buildings with available historical measurement data (labelled data), to forecast energy in physically similar buildings whose historical data is limited (some labelled data) or completely unavailable (unlabelled). For example, Fang et al. (Fang et al., 2021) address short-term energy prediction for limited historical data by introducing a hybrid deep TL strategy where long-term short memory (LSTM)-based extractor is used to infer spatiotemporal features across source and target buildings respectively. Subsequently a domain-invariant feature space is learned to bring closer the two domains using a domain adaptation neural network (DANN) and thus, the model trained on source building data can be leveraged to predict energy in a target building. Similarly, Ribiero et al. (Ribeiro et al., 2018) introduce a novel TL method based on timeseries multi-feature regression with seasonal and trend adjustments. Their approach can be used for buildings with small data by leveraging data from similar buildings with different distributions and seasonal profiles. Their work was validated using a case study involving energy prediction for a school using additional data from other schools. With a similar goal, Gao et al. (Gao et al., 2020) present a TL approach combining a sequence to sequence (seq2seq) model with a convolutional neural network (CNN), resulting in significant accuracy improvements over the use of standard LSTM given only some data in the target building. Their cross-building approach also collects data from similar buildings in a TL framework and is validated by predicting energy for three government buildings across two cities in China. Mocanu et al. (Mocanu et al., 2016) present a TL approach to predict the energy consumption at the building energy level in a Smart Grid context without the need of labelled historical data from the target building by combining a deep belief network (DBN) for feature extraction, with an extended reinforcement learning approach for knowledge transfer between buildings. The DBN estimates continuous states, which are then incorporated into the reinforcement learning methods in a continuous lower-dimensional representation of the energy consumption. The outcomes show that their model generalizes for varying time horizons and resolutions using real data thanks to the generalization power of deep belief networks.

In our approach we adopt a type of transductive TL technique called *subspace-based domain adaptation* (SDA), which differs from other TL methods by assuming the existence of a domain-invariant feature space between the source and target domains. Using a transductive approach, we leverage the facility of 'labelled data' using the Physics-based LTI SSM, by adapting the model to forecast for unobserved timesteps (unlabelled data) beyond measurement data. SDA is also referred to as *knowledge transfer* (KT) in literature. The authors are not aware of previous applications of SDA to building energy modelling. More generally, we believe domain adaptation techniques hold significant untapped benefits for building energy modelling applications including data-efficient calibration of models and cross-building generalisation, to name a few.

SDA is a type of TL approach that involves embedding/projecting source and target data into lower-dimensional vector spaces called *subspaces*. The source and target subspace are defined by the eigenvectors obtained from the ROM methods, respectively. Subsequently SDA seeks for a geometric transformation that brings the subspaces closer together in the form of a mutual subspace (Pan et al., 2011). Projecting the respective data onto a lower-dimensional subspace helps to reduce the difference between distributions between domains as much as possible while preserving important properties of the original data, such as geometric and statistical properties.

SDA is widely studied in literature with advantageous applications in fields such as computer vision (Baktashmotlagh et al., 2013; Fernando et al., 2013), natural language processing (Blitzer et al., 2006), and so forth. Most research in SDA goes toward improving the lower-dimensional embedding and/or the subspace alignment strategies with the ultimate scope of minimizing the distance between source and target. For example, Pan et al. (Pan et al., 2011) present an SDA approach to find a mutual subspace using transfer component analysis (TCA), which combines feature embedding via principal component analysis (PCA) with maximum mean discrepancy (MMD) to find an optimal shared subspace in which the distribution between source and target domain datasets is minimized. In their work, they discuss both unsupervised and semi-supervised feature extraction approaches, to dramatically reduce distance between domain distributions by projecting data onto the learned transfer components. In another approach, Huang et al. (Huang et al., 2012) adopt canonical correlation analysis (CCA) to infer a correlation subspace as a mutual representation of action models captured by different cameras for cross-view action/gesture recognition. Elhadji-Ille-Gado et al. (2017) introduce an SDA framework where source and target domains are embedded respectively into subspaces described by eigenvector matrices via an approximated singular value decomposition (SVD) method. Subsequently the source subspace eigenvectors are reoriented to align closer towards the target eigenvectors via subspace alignment, resulting in a target-aligned source subspace. In a similar approach, Fernando et al. (Fernando et al., 2013; Sebban et al., 2014) present a closed form method leading to a mapping function (transformation matrix) that aligns source subspace with target subspace vector bases, respectively derived via PCA. By aligning the bases of the subspaces, their approach is global.

### 1.4. Proposed Approach: A Physics-based Domain Adaptation Framework

Drawing inspiration from the above approaches, in this paper we present an SDA framework similar to Fernando et al., (2013) and Sebban et al., (2014). However, instead of leveraging labelled data from a source subspace, we leverage the structure of LPMs widely adopted for thermal modelling of buildings and building components, as discussed in section 1.2. More specifically, instead of a source subspace derived from labelled data, we derive the subspace directly from the eigenvectors governing the structure of the LTI SSM whose parameters are derived analytically from well-known heat transfer ODEs. On the other hand, we describe the target subspace by the eigenvectors inferred from the measurement data via unsupervised reduced order modelling (ROM) methods such as principal orthogonal decomposition (POD). Once both source and target subspaces are derived, we embed LTI SSM-simulated data (source) and observed measurement data (target) onto their respective lower-dimensional subspaces defined by the respective eigenvectors, and subsequently seek for a geometric transformation that brings the source subspace closer towards the target subspace. In dynamical systems theory, eigenvectors hold direct correspondence with the structure governing the dynamical behaviour. Thus, projecting source and target data onto their respective subspaces spanned by their respective eigenvectors, reduces the distance between the respective systems while facilitates a more data-efficient transfer. In section 5 we illustrate how the SDA approach can be used to forecast unobserved timesteps beyond available measurement data using both near and distant Physics-based approximations of the observed system.

In more detail, our Physics-based SDA approach via subspace alignment (SA) can be organized into the following steps (see Figure 1): (1) specify the physics-based model (2) centre the data, (3) infer the source subspace, (4) infer the target subspace, (5) learn the transformation map to align the source subspace with the target subspace, and finally (6) project the source data onto the target-aligned source subspace to forecast the target domain.

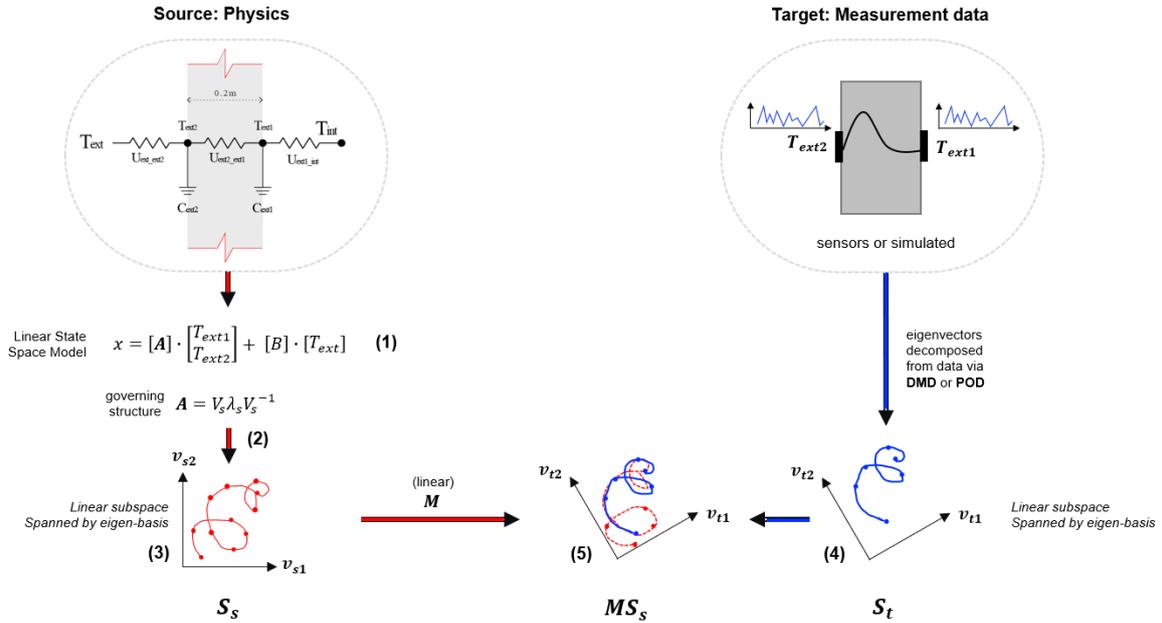

Figure 1. Overall subspace-alignment-based domain adaptation (SDA) workflow.

### 1.5. Outline

In section 2 we introduce a demonstrative scenario concerning a simple thermal system that will be carried through the paper. In section 3 we describe the specification of the physics-based state space model (step 1). Subsequently, in section 4, we describe the proposed subspace-based alignment (steps 2 to 5). Finally, in section 5, we apply the proposed approach to the demonstrative scenario and discuss the results.

## 2. Demonstrative Example

We consider a demonstrative building energy system consisting of the thermal conductance through an external wall (outlined in red in Figure 2a) of a single thermal zone. The principal energy dynamics governing the energy transfer through the external wall are represented by the RC network in Figure 2b, where $T_{ext}$, $T_{ext2}$, $T_{ext1}$ and

$T_{int}$, represent the external ambient temperature, the external surface temperature, the internal surface temperature and the internal ambient temperature, respectively. In thermal RC networks, the spatial location of the nodes is typically selected with strategic intent for the identification of hidden thermal states of interest that may not be observed or measured directly. For example, in the RC representation of walls, the node locations are typically assigned between the discretised layers composing the construction of the wall to represent the state of interaction between adjacent thermal layers thus, capture the hidden states that play a role in governing the observable dynamic behaviour. In our demonstrative example in Figure 2, nodes $T_{ext2}$ and $T_{ext1}$ were selected with the intention to capture the conductive transfer behaviour governed by the thermophysical properties of the wall and represented by the thermal resistor $R_{ext2\_ext1}$ where $R = 1/U$. The external wall separates two volumes of air where, the thermal resistors $R_{ext\_ext2}$ represents the convective heat transfer between external air and external wall surface while and $R_{ext1\_int}$ represents the convective heat transfer between the inner wall surface and internal room air. Note, that we assume a one-dimensional heat flow through the wall while we ignore the influence of solar radiation acting on the outer surface of the wall. The lumped RC network in Figure 2b is commonly labelled as the 3R2C configuration (three resistors and two capacitors) and is a well-established and validated simplified model for simulating energy behaviour of building envelopes and roofs for MPC applications in building management systems (Haves et al., 1998).

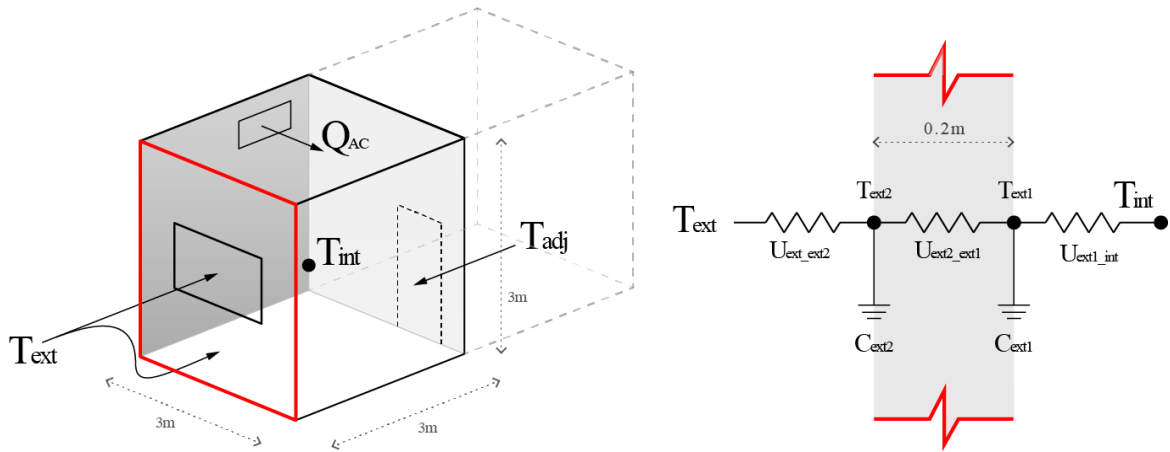

Figure 2. a) Thermal zone for context, b) thermal RC network model of 1-dimensional energy transfer through an external wall.

We can model the system of nodes algebraically using the general heat balance equation, given by:

$$\frac{dT_j}{dt} = \frac{1}{C_j}\left[\sum \frac{1}{R_k}(T_k - T_j) + \sum Q_j\right] \quad (1)$$

Where, $j$ and $k$ represent two adjacent thermal nodes; $C_j$ represents the thermal storage capacity at node $j$; $R$ represents the thermal resistance connecting $j$ and $k$; $Q$ represent heat gains from external sources such as lighting, and equipment. For the simple wall scenario, we assume a homogenous 1D energy conduction through the wall while ignore heat gains $Q$ from the zone adjacent to the inner surface of the wall. Using Equation (1), we can derive the respective energy balance equations for each node $[T_{ext1}, T_{ext2}]$ in the 1D wall scenario as follows:

$$\frac{dT_{ext1}}{dt} = \frac{1}{C_{ext1}}\left[\frac{1}{R_{ext2_{ext1}}}(T_{ext1} - T_{ext2})\right] \quad (2)$$

$$\frac{dT_{ext2}}{dt} = \frac{1}{C_{ext2}}\left[\frac{1}{R_{ext2\_ext1}}(T_{ext1} - T_{ext2}) + \frac{1}{R_{ext\_ext2}}(T_{ext2} - T_{ext})\right] \quad (3)$$

The $R$ and $C$ parameters can either be identified using various black-box system identification optimization methods (Candanedo et al., 2013; Madsen, 1995; Sarkar et al., 2021) from numerically--simulated or empirically-measured data or can be derived directly from first principles. Given our aim to leverage Physics-derived knowledge of phenomenon, we derive $R$ and $C$ directly from first principles of heat storage capacitance, convective resistance, and conductive resistance using well-established mechanisms Equations (4), (5) and (6).

$$C = Cp\,\rho\,V \tag{4}$$

$$R_{conv} = 1/{h\,A} \tag{5}$$

$$R_{cond} = kA/{w/2} \tag{6}$$

Table 1. Assigned Physical and thermodynamic properties for the 1D wall scenario

| Genre | Property, notation | Value |
|---|---|---|
| Physical properties | Wall layer volume, $V$ | 1.8 $m^3$ |
| | Wall thickness, $w$ | 0.2 $m$ |
| Thermophysical brick properties | Brick conductivity, $k$ | 0.72 $W/mK$ |
| | Brick density, $\rho$ | 1920 $kg/m^3$ |
| | Brick specific heat capacity, $Cp$ | 780 $J/kgK$ |
| Convection coefficients | Indoor convection, $h$ | 8 $W/m^2K$ |
| | Outdoor convection, $h$ | 25 $W/m^2K$ |
| Thermophysical air properties | Air density, $\rho$ | 1.2 $kg/m^3$ |
| | Air specific heat capacity, $Cp$ | 100 $J/kgK$ |

## 3. Specification of the Physics-based Model (Step1)

The first step in our SDA approach is the specification of the Physics-based model. In contrast to standard SDA approaches discussed earlier, in our approach we assume a Physics-based LPM as the source domain. More specifically we assume a Physics-derived LTI SSM which offers a standardised representation of dynamical systems and has been shown to capture governing behaviour of energy in buildings accurately even when parameters are lumped (Goyal et al., n.d.; Goyal & Barooah, 2012a). Furthermore, lumped SSMs for building thermal network representations can be relatively straightforward to implement.

### 3.1. LTI SSM

We can further represent the coupled ODEs in Equations (2) and (3) in state space form which is a more compact format to represent the mechanistic structure of the dynamics. State space models (SSM) are a standardised representation to describe the dynamics of any system which can be described by a set of ODEs or partial differential equations (PDEs) (Hendricks et al., 2008). In general, SSMs consist of two matrix equations: the state equation (Equation (7)) describes how the inputs influence the states of the system while the output equation (Equation (8)) describes how the states of the system and the external factors (inputs) directly influence the observed output of the system. In this paper we focus on discrete-time LTI SSMs which have the following form:

$$x(t) = [A] \cdot x(t-1) + [B] \cdot u(t-1) \tag{7}$$

$$y(t) = [C] \cdot x(t-1) + [D] \cdot u(t-1) \tag{8}$$

LTI implies that the coefficient parameters do not vary with time. From here on it is implied that SSM refers to LTI SSM. SSMs are written in terms of three vectors: a state vector $x$ with $n$ elements; an input vector $u$ with $p$ elements; and an output vector $y$ with $q$ elements. Such a state-space formulation is equivalent to the well-known finite difference method. We find that the SSM formulation provides a compact and standardised format to preserve the interpretable structure of governing dynamics, with focus for knowledge transfer purposes described in section 3.

When representing RC thermal networks as SSMs, we can adopt the temperatures at the capacitance nodes in the RC network as the state variables in the state vector $x$, to maintain the same intuitive interpretation (Candanedo et al., 2013). In our example wall scenario, the temperature on the internal surface of the wall is our variable of interest and thus, we consider state $T_{ext1}$ as the output variable $y$ while $T_{ext}$ as the control input $u$, as follows:

$$x = \begin{bmatrix} T_{ext1} \\ T_{ext2} \end{bmatrix}, u = [T_{ext}], y = [T_{ext1}] \tag{9}$$

These vectors are linked by four coefficient matrices, $A$ $(n \times n)$, $B$ $(n \times p)$, $C$ $(q \times n)$ and $D$ $(q \times p)$ that capture the dynamic temporal behaviour between states $x$, inputs $u$ and output $y$. More specifically, $A$ is referred to as the dynamics operator which contains the dynamical characteristics to advance the state matrix $x$ from in Equation (7) from $x(t)$ to $x(t+1)$, for time step $dt$, for any arbitrary $x(t)$. $B$ determines how the influences from control signals e.g., the outdoor air temperature, affect the states e.g., the inner and outer wall surface temperatures. The zero values in $B$ indicate that the outdoor temperature $T_{ext}$ influences the internal wall surface temperature $T_{ext1}$ by conduction through the wall, not directly. The coefficient values of $A, B, C, D$ are determined directly using the heat balance equations at each node. Thus, the coefficients for the wall example are determined via Equations (2) and (3) and as follows:

$$A = \begin{bmatrix} \frac{-U_{ext1,ext2}}{C_{ext1}} & \frac{U_{ext1,ext2}}{C_{ext1}} \\ \frac{U_{ext1,ext2}}{C_{ext2}} & \frac{-U_{ext1,ext2}-U_{ext2,ext}}{C_{ext2}} \end{bmatrix}, B = \begin{bmatrix} 0 \\ \frac{U_{ext2\ ext}}{C_{ext2}} \end{bmatrix}, C = \begin{bmatrix} 1 \\ 0 \end{bmatrix}, D = [0] \tag{10}$$

### 3.2. Structure of the Governing State Space

A state-space representation allows for a more geometric understanding of dynamical systems. The operator $A$ can be decomposed into its fundamental components by:

$$AV = V\lambda \tag{11}$$

where, $V$ and $\lambda$ are the eigenvectors $v_1, v_2$ and eigenvalues $\lambda_1, \lambda_2$ of describing the dynamics of $A$, respectively. The system $A$ can then be solved for any timestep $dt$ and initial condition $x(t_0)$ by solving the matrix exponential method as follows:

$$\Phi_A = e^{dtA} = Ve^{dt\lambda}V^{-1} \tag{12}$$

$$\dot{x}(t) = e^{tA}x(t_0) \tag{13}$$

where $\Phi_A$ is referred to as the state transition matrix; the matrix that updates the state vector from one timestep to the next. The eigenvalues lie on the diagonal of the A matrix. Thus, when computing the matrix exponential, we are scaling/transforming the eigenvalues to a given timescale, $dt$.

We demonstrate the generalizability and interpretability of this physics-derived model using portrait analysis, which is a qualitative analysis method from dynamical system theory to visualize the structure and solutions of dynamical systems geometrically, as trajectories in a vector field. Using portrait analysis, we aim to illustrate how $V$ and $\lambda$ govern the global physical behaviour of dynamics, in our case, thermal conduction through the façade wall of a building.

In portrait analysis, instead of viewing the state variables $(T_{ext1}, T_{ext2})$ independently as functions of time (as a timeseries in Figure 3a), we can view their values as coordinates of a point in a vector space bound by $T_{ext1}, T_{ext2}$ called the *state space* (Figure 3b). Thus, a point in the state space represents the complete state of the system at any time $t$. As the system evolves over time, the point will trace out a curve in the state space, referred to as a *trajectory*. Figure 2b illustrates the traced trajectory in the state space for 300 hourly timesteps for an exemplary scenario where initial internal ($T_{ext1}$) and external ($T_{ext2}$) wall surface temperature conditions (time $t = 0$) started at 10.73°C and 10.82°C, respectively. This exemplary initial state scenario could correspond to a situation where an indoor zone experiences heat loss after being recently ventilated (e.g., due to open window for long period of time) causing the internal zone air temperature to drop closer towards the outdoor temperature.

The behaviour of the trajectory observed in Figure 3b can be described as a linear combination of thermal conduction dynamics between $T_{ext1}$ and $T_{ext2}$, and the influence of the ambient temperature $T_{ext}$, as described by the state equation in Equation (7). In terms of control systems theory, $T_{ext}$ can be described as a control forcing signal. Therefore, if we decouple the $T_{ext}$ from $T_{ext1}$ and $T_{ext2}$, we reveal the trajectory traced by the conduction dynamics alone (blue curve in Figure 4a.) We do this by fixing the ambient outdoor temperature to a constant value (0°C), i.e., we consider $x(t + 1) = Ax$.

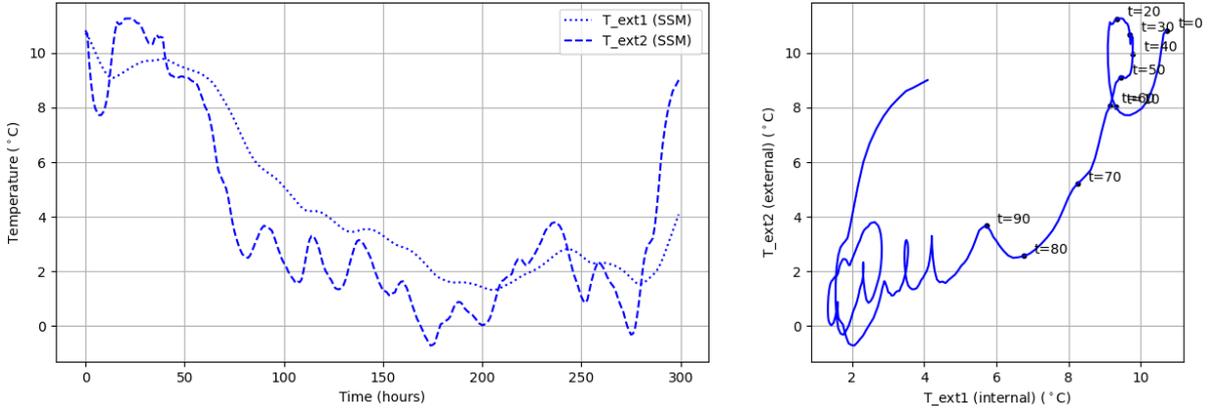

Figure 3. a) SSM-generated timeseries measurement data plotted as a function of time, b) SSM-generated timeseries measurement data plotted in the state space.

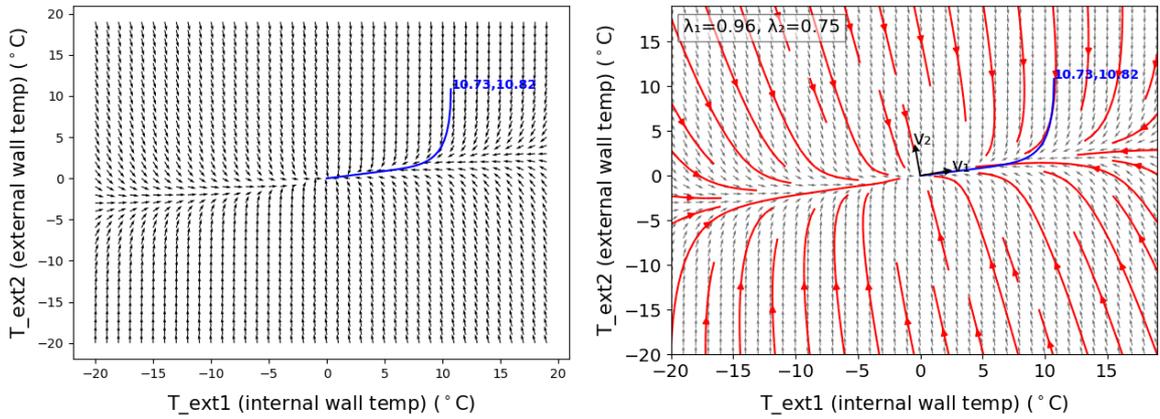

Figure 4. a) State space vector field, b) Phase portrait of the state space.

The geometry of the decoupled trajectory describes how the full state of the system approaches steady state conditions, i.e., thermal equilibrium. We can interpret the geometry as follows: shortly after the room was ventilated (initial conditions at time $t = 0$), the external surface of the wall ($T_{ext2}$) experiences a sudden drop in temperature due to a decrease in outdoor ambient temperature. Note how the internal surface temperature ($T_{ext1}$) starts to drop at a much slower rate due to the thermal resistance of the wall. Finally, the internal surface temperature drops towards zero very rapidly once the thermal capacity of the wall is reached. In other words, the path traced by the trajectory can be interpreted as the dynamical behaviour of conduction given the specified thermophysical properties of the wall in Table 1.

The velocity and path of the traced trajectory are governed by an underlying vector field which describes the global behaviour of any trajectory in the state space. The vector field can be described fully by the governing dynamics matrix $A$ in the state space model from Equation 7. More specifically, the velocity and direction of any coordinate $T_{ext1}, T_{ext2}$ in the state space vector field is a function of the eigenvectors $v_1, v_2$ and eigenvalues $\lambda_1, \lambda_2$ composing $A$. In fact, any point in the state space $T = (T_{ext1}, T_{ext2})$ (representing a thermal scenario) can be described as a linear combination of the eigenvectors. The velocity in a thermal state space represents the rate of change of the temperature states over time and is dictated by the eigenvalues also interpreted as 'thermal eigenvalues' (Hyun & Wang, 2019; Madsen, 1995). Thus, by generating solution trajectories for arbitrary initial state combinations $T(t_0) = [T_{ext1}(t_0), T_{ext2}(t_0)]$ we reveal the structural geometry of the dynamics governing the thermal behaviour of the wall in Figure 4b (also known as a phase portrait). It is this innate initial condition generalizability that we aim to leverage for the knowledge transfer (and this level of physical interpretability for the wider research goal).

Note that phase portraits are more typically used to plot the state space of continuous-time systems however, the plots in this paper are for discrete-time systems as a result of the post matrix exponential of the dynamics matrix $A$ in Equation (12).

Figure 5a) and Figure 5b) illustrate the state trajectories for heat transfer through thicker walls, of 0.6m and 1.5m respectively. On qualitative comparison with Figure 3a, we can instantly note a change in the geometry of the vector field and consequently the trajectories, governed by a rotational transformation in the respective orthogonal eigenvector pairs. As the wall gets thicker and the combined U-value decreases, the gradient of heat transfer decreases. When comparing the state space geometry for different wall systems, e.g., varying thicknesses, we can instantly observe a mappable/interpolative similarity. The geometric similarity of these independent but related systems begs the question if mechanistic knowledge can be transferred across a range of thermal systems if represented geometrically as such. Furthermore, Figures 15 to 17 in Appendix D illustrate the geometric differences between dynamics of a system derived from first principles and dynamics derived from data dynamics using spectral decomposition techniques. This view opens up a geometric way to approach/develop geometry-based calibration techniques. This is the core motivation behind our work and why we look towards DA as a means to transfer mechanistic knowledge.

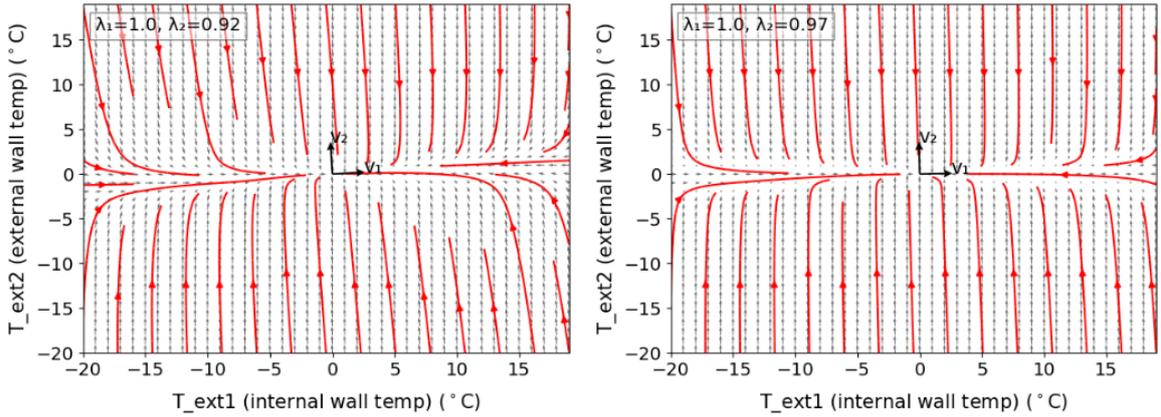

Figure 5. a) Heat transfer through 0.6m thick wall, b) Heat transfer through 1.5m thick wall.

## 4. Physics-based Domain Adaptation

### 4.1. Source and Target Data Pre-processing (Step 2)

We start by generating a source dataset $X_s \in \mathbb{R}^{n \times z}$, using the Physics-based SSM which is computationally inexpensive to run. On the other hand, we assume availability of a set of measurement data $X_t \in \mathbb{R}^{n \times z}$ which for this paper we generate synthetically (from an EnergyPlus model of the same system) as a representation of empirically observed building measurement data. It is our goal to learn a matrix that geometrically aligns $X_s$ with $X_t$. Note that in this approach we assume pairwise correspondence between source and target data points, where both are 'generated' under the same input conditions.

Subsequently, $X_s$ and $X_t$ are split according to a training/testing ratio ($l/m$). The target training set is used to derive the target subspace eigenvectors $V_t$ via one of the ROM methods discussed, and subsequently also used to learn the transformation matrix $M$ in Step 4. On the other hand, the target test set is used to validate the forecasted data. The size of $V_t$ used to train the ROM may be set independently from the size of $V_t$ used to learn the transformation.

Prior to subspace embedding, the source and target data are centred along the zero mean as specified by Equation (14). Centring the data is an important step and critical for subspace alignment as it brings closer the source and target distributions before alignment. The training set and testing set are centred independently to avoid prediction bias due to data leakage from training information to testing.

$$\bar{X} = \frac{1}{n}\sum_{j=1}^{n} X_j = 0 \tag{14}$$

### 4.2. Infer Source Subspace (Step 3)

In contrast to standard SDA approaches, we derive the basis of the source subspace as eigenvectors $V_s$ inferred directly from the Physics-derived low-rank SSM representation of the energy system in step 1, via eigendecomposition of its system matrix $A$. The decomposition is described in terms of the eigenvalue problem in Equation (15) where $\lambda_s$ is a diagonal matrix containing the eigenvalues. In typical SDA applications the eigenvectors are derived directly from source data via ROM methods and represent the principal directions in which the variance spans. In our case, the eigenvectors (and eigenvalues) hold direct connotation with the governing structure of the state space as illustrated in section 2 and thus, aid to improve generalizability of the learned transformation when embedding the measurement data into (lower dimensions) subspaces. It is worthy to note that the $A$ derived from the thermal RC network is symmetric implying that the eigenvectors are linearly independent and orthogonal.

$$A = V_s \lambda_s V_s^{-1} \tag{15}$$

### 4.3. Infer Target Subspace (Step 4)

In our approach we derive the vector basis of the target subspace $V_t$, directly from the available measurement data $X_t$ via ROM. In this paper, we adopt principal orthogonal decomposition (POD) as the ROM method of choice.

POD is a linear and unsupervised ROM method widely employed in various fields including signal analysis and pattern recognition. POD is equivalent to the well-known principal component analysis (PCA) in that they both employ singular value decomposition SVD to decompose the matrix into orthogonal eigenvectors, aka orthogonal modes. These modes are then used to project the original dataset into an optimal lower-dimensional[1] approximation. When applied to fields pertaining to dynamic behaviour, the eigenvectors obtained are interpreted as characteristic spatial structure also known as modes, which refer to the characteristic structure of the source system, especially in flow dynamics.

In typical SA applications, further method extensions are implemented to account for deriving the ideal dimensionality of source and target subspaces when applying ROM to the source and target data. The same does not apply to our proposed approach, where both source and target eigenvectors correspond directly to the states of the system thus, already exist in the governing dimensionality.

### 4.4. Subspace Alignment (Step 5)

The focus of most SDA methods is to learn an optimal matrix transformation operator/s that aligns the source and target subspaces. We explore two alternative sub-methods for subspace alignment: standard SA via minimisation of the Bergman divergence perspective (Fernando et al., 2013) and Procrustes-based SA (C. Wang & Mahadevan, 2008). A summary of both algorithms used is given by Algorithm 1, and Algorithm 2 and are further discussed in more detail here:

The classic SA method is the standard DA alignment approach where a linear transformation that maps source subspace to the target one is determined. This is achieved by aligning the basis vectors using a transformation matrix $M$ from $X_s$ to $X_t$ ($M \in \mathbb{R}^{d \times d}$) which is learned by minimizing the Bergman matrix divergence in Equation (16) and Equation (17), where $\|.\|_F^2$ is the Frobenius norm. The Frobenius norm is invariant to orthonormal operations and thus can be rewritten as the objective function in Equation (19) as suggested by Fernando et al., (2013); Sebban et al., (2014).

$$F(M) = \|X_s M - X_t\|_F^2 \tag{16}$$

$$M^* = argmin_M F(M) \tag{17}$$

---

[1] While in our demonstrative application in section 5 we deal with a simple system that does not require order reduction, our approach applies with reduced-order states.

$$M^* = X_s^T X_t \tag{18}$$

$$F(M) = \|X_s M - X_t\|_F^2 \tag{19}$$

In the second alignment approach we adopt Procrustes analysis, which is a classic statistical method typically used for shape analysis and image registration of 2D/3D data, which seeks isotropic dilation and the rigid translation, reflection and rotation needed to best match one data structure to another. More specifically, Procrustes alignment removes the translational, rotational, and scaling components so that the optimal alignment between source-target instance pairs is found.

Instead of finding a single transformation matrix $M$, Procrustes manifold alignment (Perron et al., 2021; C. Wang & Mahadevan, 2008) seeks to align source and target subspaces using the rotation matrix $r \in \mathbb{R}^{n \times n}$ and scaling factor $s$, found via Procrustes analysis on the embedded data. First, we project SSM-generated data onto the source eigenvectors, and the target measurement data onto the target eigenvectors,

$$\tilde{X}_s = X_s V_s^T \, , \, \tilde{X}_t = X_t V_t^T \tag{20}$$

Subsequently, we align the two using ordinary Procrustes via the singular value decomposition of the embedded datasets:

$$U\Sigma V^T = SVD(\tilde{X}_t^T \tilde{X}_s) \tag{21}$$

$$r = UV^T \tag{22}$$

$$s = \frac{trace(\Sigma)}{trace(\tilde{X}_s (\tilde{X}_s)^T)} \tag{23}$$

where $U \in \mathbb{R}^{n \times n}$ and $V \in \mathbb{R}^{n \times n}$ are matrices containing the left and right singular vectors and $\Sigma \in \mathbb{R}^{n \times 1}$ is a diagonal matrix containing the singular values on its diagonal. Note that in this implementation, we ignore the translation term during Procrustes since given that both $\tilde{X}_s$ and $\tilde{X}_t$ are centered to zero mean prior to subspace embedding. Finally, we reconstruct the original training data by applying $r$ and $s$ to the embedded source data (Equation (24)) and lifting it into the original target space (Equation (25)). With this in place, we can forecast for new inputs beyond the available target data.

$$\tilde{X}_a = sr\tilde{X}_s \tag{24}$$

$$NX_a = V_s \tilde{X}_a^T \tag{25}$$

**Algorithm 1. SA-based DA via Bergman divergence**

Inputs: SSM generated data $X_s$, measurement data (simulated) $X_t$

Output: Forecasted target data $X_{\hat{t}}$

1. Infer source eigenvectors $V_s$ from the SSM by decomposing $A$ where $A = V_s \lambda_s V_s^{-1}$
2. Infer target eigenvectors $V_t$ from measurement data via POD
3. Obtain optimal $M$ by aligning $V_s$ towards $V_t$ via minimisation of the Bergman divergence where $M = argmin_M(\|V_s M - V_t\|_F^2)$
4. Apply $M$ to obtain the basis of the new target-aligned subspace $V_a$ where $V_a = V_s M$
5. Project source data $X_s$ onto new target-aligned subspace $V_a$ where $\tilde{X}_a = X_s V_a^T$
6. Reconstruct aligned source data in target domain where $X_a = \tilde{X}_a V_t$
7. Forecast $X_{\hat{t}}$ for new inputs by repeating steps 3, 6 and 7

**Algorithm 2. SA-based DA via Procrustes analysis**

Inputs: SSM generated data $X_s$, measurement data (simulated) $X_t$

Output: Forecasted target data $X_{\hat{t}}$

1. Infer source eigenvectors $V_s$ via decomposition of state transition matrix $A$ into $V_s \lambda_s V_s^{-1}$
2. Infer target eigenvectors $V_t$ from data via POD
3. Project source data $X_s$ onto source subspace $V_s$ where $\tilde{X}_s = X_s V_s^T$
4. Project target data $X_t$ onto target subspace $V_t$ where $\tilde{X}_t = X_t V_t^T$
5. Infer rotation matrix $r$, scaling factor $s$ and translation factor $t$ via Procrustes analysis where $r, s, t = argmin(\|X_s - kX_t r\|_F)$
6. Apply $r, s, t$ transformations to $\tilde{X}_s$ via $\tilde{X}_a = rs\tilde{X}_s + t$
7. Reconstruct aligned target data via $X_a = V_t \tilde{X}_a^T$
8. Forecast $X_{\hat{t}}$ for new inputs by repeating steps 3, 6 and 7

### 4.6. Error metrics

We quantify the performance of the subspace alignment-based DA by comparing its prediction accuracy for out of sample inputs, to that of the target data which we consider to be the 'true' observed data. More specifically, we compute the coefficient of variation of root-mean squared error (RMSE) and the normalised mean bias error (NMBE) as follows:

$$CV(RMSE)(X) = \frac{1}{\bar{X}} \sqrt{\frac{1}{n} \sum_{i=0}^{n-1} (X_i - \hat{X}_i)^2} \tag{26}$$

$$NMBE(X) = \frac{\sum_{i=0}^{n-1}(\hat{X}_i - X_i)}{\sum_{i=0}^{n-1}(X_i)} \tag{27}$$

We adopt the guidelines set by the building energy modelling and forecasting community (ASHRAE, 2014) to determine the validity of the RMSE and NMBE results for energy forecasting. For hourly model calibration, it is recommended that maximum allowed NMBE is capped at 10% while the CV(RMSE) at 30%.

### 5. Demonstrative application

As aforementioned, we illustrate two application scenarios with the demonstrative wall example: 1) calibration of Physics-simulated data to measurement data, 2) reuse of physics-based model for a variation of wall types undergoing similar phenomenon. For each scenario we plot and report pre-and post-alignment results via CVRMSE and NMBE measures for timeseries reconstruction and forecasting.

### 5.1. Transfer across equivalent systems (calibration)

We first simulate 7000 hours of synthetic measurement temperature data for a 0.2m wall scenario using a high-fidelity simulation model of the thermal zone in Figure 2 using EnergyPlus, which is a widely simulation software used by the building energy community for thermal modelling and simulation. The goal is to calibrate data simulated using a physics-derived SSM model of the same wall scenario (source), closer towards the simulated measurement data (target).

For this study we select varying training sets from the simulated high-fidelity data: 2000 hours, 4000 hours, 6000 hours. During the simulation, we record the temperature on the inner surface $T_{ext1}$ and exterior surface $T_{ext2}$ of the wall at an hourly timestep. Our goal in this exercise is to forecast beyond the measurement data assigned as training ($T_{ext1}target, T_{ext2}target$) by leveraging response data ($T_{ext1}source, T_{ext2}source$) generated from a governing Physics-based description (SSM) of the same 0.2m thick wall scenario, and which is computationally inexpensive to obtain. Note that for clarity, throughout this study we illustrate only the alignment and forecast for the temperature of the inner surface of the wall $T_{ext1}$ since it is significantly influenced by the conductive dynamics through the wall and thus, of main interest.

We first specify the elements of the system matrix $A$ by substituting the thermophysical values for a wall of 0.2m thickness in Table 1 into Equation (9). Subsequently, we obtain the state transition matrix $\Phi_A$ for an hourly timestep $dt$ by solving $e^{dtA}$ (Equation (29)). Here, we assume an hourly timestep $dt = 3600s$. Given $\Phi_A$, we can obtain the eigenvectors $V_{s1}, V_{s2}$ (specified in Appendix A, Table 5), via the decomposition in Equation (12).

$$A = \begin{bmatrix} -1.2019e-05 & 1.2019e-05 \\ 1.2019e-05 & -7.879e-05 \end{bmatrix} \qquad (28)$$

$$\Phi_A = e^{dtA} = \begin{bmatrix} 0.95848 & -0.03684 \\ -0.03684 & 0.75379 \end{bmatrix} \qquad (29)$$

Once we compute the basis eigenvectors of the source subspace, we proceed to extract the basis eigenvectors $V_{t1}, V_{t1}$ of the target subspace from the available measurement data using data-driven ROM –here, we use POD to infer the basis of the target subspace. The target eigenvectors derived via POD are specified in

Table 7 in Appendix A.

Now that we have obtained the vector bases for both source and target subspaces, we first project the data generated by the SSM ($T_{ext1} source$, $T_{ext2} source$) onto the source subspace (onto $V_{s1}, V_{s2}$), then project the measurement data ($T_{ext1} target$, $T_{ext2} target$) onto the target subspace (onto $V_{t1}, V_{t2}$), and subsequently proceed to learn a transformation map $M$ by aligning the projected source data closer towards the projected target data using the a) standard SA via Bergman divergence minimisation and b) Procrustes-based SA, described earlier. To reiterate, with a) we learn a rotational matrix only while with b) we learn a rotational matrix, together with scaling and translation factors. Once obtained, the transformation map $M$ is first applied to reconstruct the target measurement data used for training by aligning the source data to the target data, and subsequently used to geometrically transform new data generated by the Physics-derived SSM into target-aligned data to forecast for 1000 hours beyond the target measurement data. The CV(RMSE) forecasting results via the standard SA Bergman divergence and Procrustes SA approaches are tabulated in Table 2 and Table 4, respectively to aid comparison between the two alignment approaches for a variety of transfer scenarios. The standard SA Bergmann approach performs poorly at the get-go as can be noted by the CV(RMSE) results. For this reason, in this section we discuss and illustrate plots for Procrustes-based alignment only.

Table 2. CV(RMS) accuracy for forecasting 1000 hours using Standard SA via Bergmann divergence.

|  | Standard SA (Bergmann divergence) | |
| --- | --- | --- |
| Training size (hrs) | 0 | |
| Thickness ($metres$) | $0.2_{ssm} \rightarrow 0.2_{true}$ | $0.8_{ssm} \rightarrow 0.8_{true}$ |
| $CVRMS$ SSM (°C) | 8.74% | 6.34% |
| $CVRMS_{POD}$ aligned (°C) | 23.99% | 10.66% |

Figure 6 illustrates the reconstruction of the internal wall surface temperature $T_{ext1}$ via Procrustes-based alignment of the Physics-derived $T_{ext1}$ data (red line). Subsequently, the learned transformation was applied to forecast a further 1000 hours (dotted red line). We note an overall post-alignment improvement in the forecasting CV(RMSE) and NMBE values. We observed that while the Physics-derived SSM already approximates the measured system quite well with a CV(RMSE) and NMBE within a desirable bound (≤30% and ≤10%, respectively), our SDA approach improved further these forecasting accuracy values by 1.56% and -3.57%. In the following figures we also plot the reconstruction and forecasting error (bottom) as a function of localised difference between the aligned and target datapoints (error_postaligned). We compare this with the localised difference between the output from the Physics-derived SSM and the target datapoints (error_prealigned).

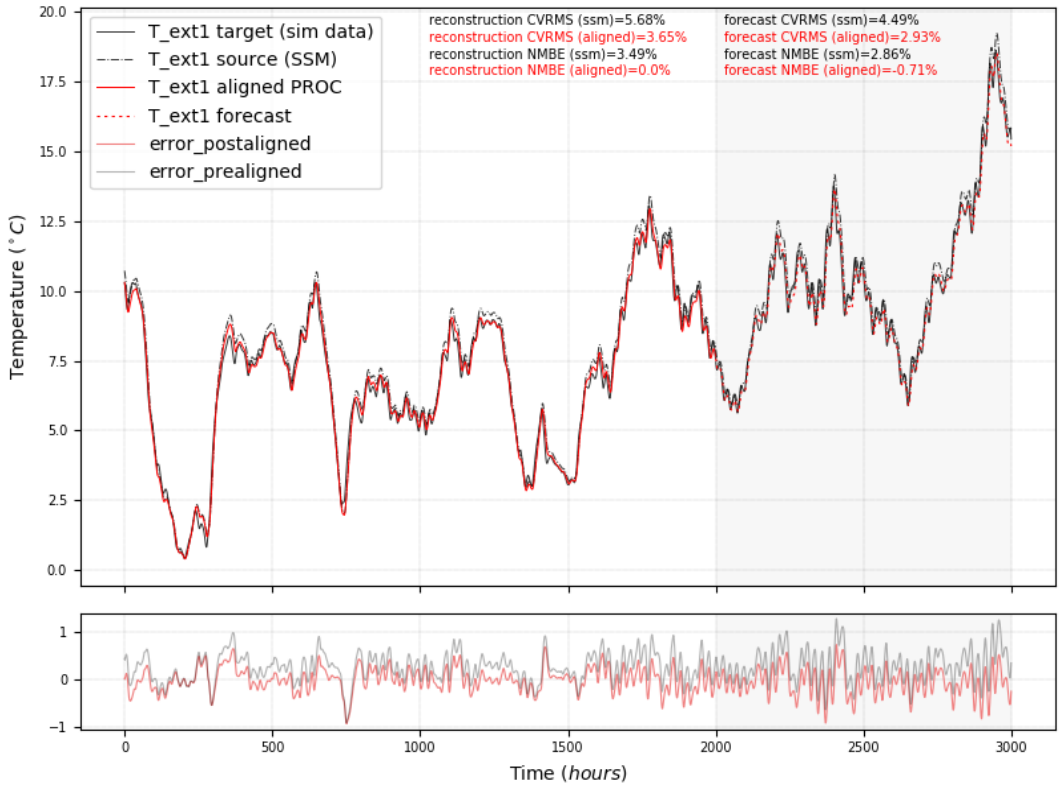

Figure 6. Calibration scenario: 0.2m wall SSM (source) to 0.2m wall data (target). The target subspace was derived via POD (orthogonal eigenvectors).

We repeated the calibration study for a thicker wall (0.8m) using Procrustes-based SA on 2000 hours training data. From Table 3 we note that the forecasting accuracy of the aligned (transformed) source data in the 0.8m wall calibration scenario is 4.49% worse than the original Physics-derived SSM (6.34%). We observed improvement for this case when increasing the training data size.

Table 3. CV(RMS) accuracy for forecasting 1000 hours using Procrustes-based SA for varying training size (2000 hours, 4000 hours, 6000 hours).

|  | Procrustes-based SA | | Procrustes-based SA | | Procrustes-based SA | |
| --- | --- | --- | --- | --- | --- | --- |
| Training size (hrs) | 2000 | | 4000 | | 6000 | |
| Thickness (*metres*) | $0.2_{ssm} \rightarrow 0.2_{true}$ | $0.8_{ssm} \rightarrow 0.8_{true}$ | $0.2_{ssm} \rightarrow 0.2_{true}$ | $0.8_{ssm} \rightarrow 0.8_{true}$ | $0.2_{ssm} \rightarrow 0.2_{true}$ | $0.8_{ssm} \rightarrow 0.8_{true}$ |
| $CVRMS$ SSM | 4.49% | 7.53% | 2.75% | 3.57% | 2.45% | 12.00% |
| $CVRMS_{POD}$ aligned | 2.93% | 5.05% | 2.25% | 6.07% | 1.76% | 9.81% |

### 5.2. Transfer across different but related systems (reuse)

We hypothesise that building energy systems governed by similar phenomena have geometrically similar state space structures implying that their governing subspaces live in a mutual latent space. This suggests that mechanistic descriptions of energy transfer through one wall could be used to describe and forecast the energy transfer through that of a wall with different geometric and material properties, undergoing similar energy transfer phenomena. In this context, we demonstrate how the SDA approach can be used to leverage the low-rank Physics-derived SSM of one wall to predict and forecast the internal wall surface temperatures of observed walls with varying thickness and material properties, by learning a subspace alignment transformation between Physics and given wall surface temperature measurement data.

Figure 7 illustrates the adaptation of a Physics-derived SSM for a 0.6m wall to forecast beyond data observed from a 0.2m wall via the POD-derived target subspace. The respective source and target eigenvectors can be found in Table 6 in Appendix A. We can immediately observe how the transformation learned between the Physics-derived data and the target data has successfully been captured by its ability to reconstruct the training set (red line) when applied to the source data and subsequently to forecast a further 1000hours (dotted red line). We utilise the CV(RMSE) and NMBE values to compare the prediction improvement while ensure we are still in desirable bounds typically adopted for building energy model calibration (≤30% and ≤10%, respectively). The improvement can also be noted by observing the difference in pre and post alignment error timeseries in the bottom plots.

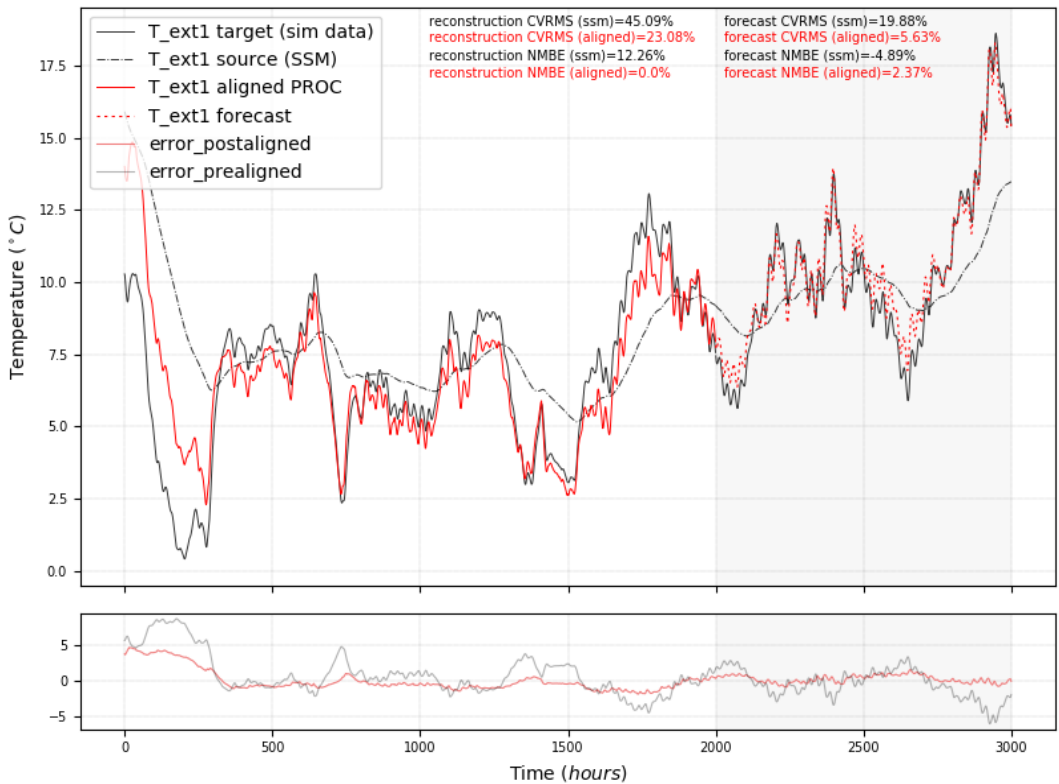

Figure 7. Cross-domain generalization scenario: 0.6m wall SSM (source) to 0.2m wall data (target). The target subspace was derived via POD (orthogonal eigenvectors).

On the other hand, Figure 8 illustrates the adaptation of a Physics-derived SSM for a 0.2m wall to forecast beyond data observed from a 0.6m wall. While we note forecasting improvement indicating a closer match to the target measurement data, we observe that the transformation via Procrustes alignment does not dampen the noisier source signal. In section 5.3 we take a closer look to interpret why this occurs.

Lastly, Figure 9 illustrates the adaptation of the SSM for a red brick wall of 0.8m thickness, 0.72 $W/mK$ conductivity, 1920 $kg/m^3$ material density, and 780 $J/kgK$ specific heat capacity, to forecast beyond temperature data measured from a concrete wall of 0.8m thickness, 1.3 $W/mK$ conductivity, 2240 $kg/m^3$ material density, and 840 $J/kgK$ specific heat capacity. The respective source and target eigenvectors can be found in Table 7 in Appendix A. We can immediately observe that the transformation learned between the SSM and target data is able to forecast the internal surface temperature of the concrete wall for unseen external temperature inputs. This can be noted by the improvements on CV(RMSE) and NMBE values.

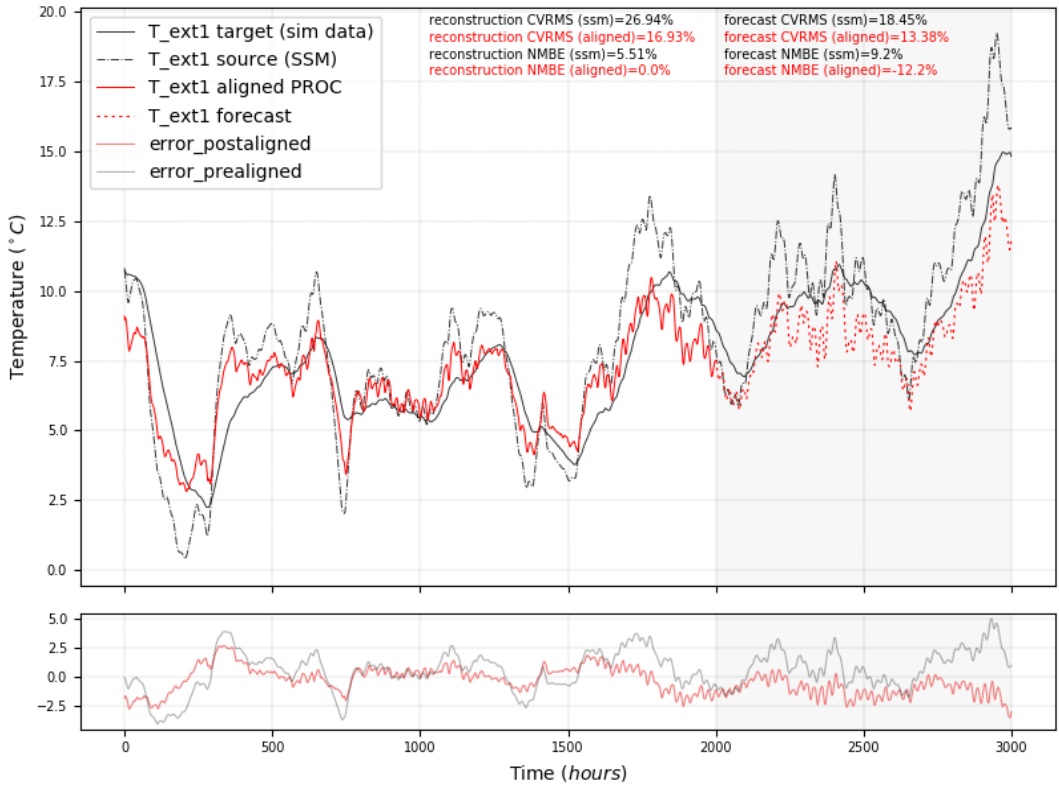

Figure 8. Cross-domain generalization scenario: 0.2m wall SSM (source) to 0.6m wall data (target). The target subspace was derived via POD (orthogonal eigenvectors).

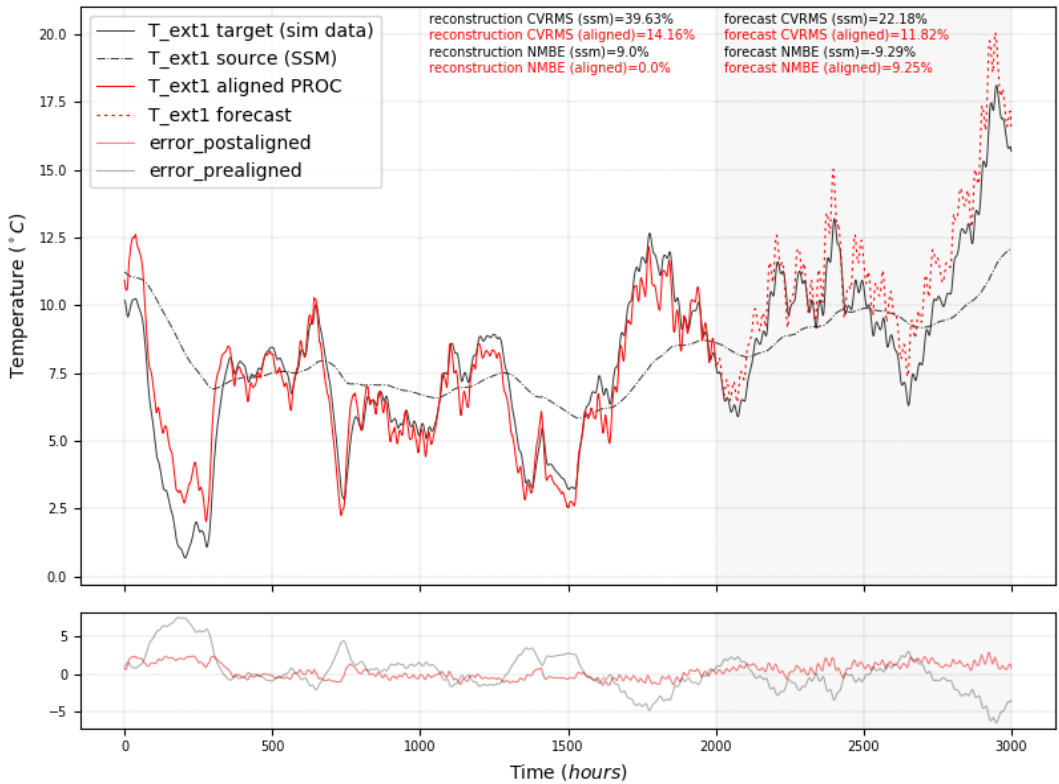

Figure 9. Cross-domain generalization scenario: 0.8m red brick wall SSM (source) to 0.3m concrete wall data (target). The target subspace was derived via POD (non-orthogonal eigenvectors).

## 5.3. Discussion

The standard SA method via Bergmann divergence minimisation does not depend on data because it acts only on the subspaces while Procrustes-based SA acts directly on the local geometry of the projected source and target data. The former only accounts for rotational transformational between source and target subspaces while the latter yields a rotation matrix, skewing, translation, and reflection factors. In section 5.1 we observed that rotation only is not sufficient to completely describe geometric mappings between subspace-embedded source and target data, particularly in cases where the source and target eigenvectors are in a mirrored orientation from each other. Thus, while we acknowledge that Procrustes-based SA is data-dependant, it is suitable for capturing geometric transformations between Physics-derived and measurement timeseries signals, even when transferring across different but related wall-types (Table 4).

Table 4. CV(RMS) accuracy for forecasting 1000 hours using cross-domain Procrustes-based SA for varying wall thicknesses only.

|  | Procrustes-based SA | | | | | |
| --- | --- | --- | --- | --- | --- | --- |
| Training size (hrs) | 2000 | | | | | |
| Thickness ($m$etres) | $0.2_{ssm} \rightarrow 0.6_{true}$ | $0.2_{ssm} \rightarrow 0.8_{true}$ | $0.2_{ssm} \rightarrow 0.9_{true}$ | $0.6_{ssm} \rightarrow 0.2_{true}$ | $0.8_{ssm} \rightarrow 0.2_{true}$ | $0.9_{ssm} \rightarrow 0.2_{true}$ |
| $CVRMS$ SSM | 18.45% | 23.02% | 23.59% | 19.88% | 23.5% | 25.09% |
| $CVRMS_{POD}$ aligned | 13.38% | 16.75% | 16.82% | 5.63% | 5.99% | 11.85% |

We observed that when lifting the target-aligned data from the embedded space to the target space, the source-target alignment can be compromised in cases when the basis of the target-aligned subspace does not align well with the basis of the target subspace. For example, in the cross-domain alignment example ($0.2_{ssm} \rightarrow 0.9_{true}$) illustrated in Figure 10, top, we instantly note that Procrustes successfully aligns the source with target signal in the embedded space (left) but the alignment is subsequently compromised when lifting into the target subspace (right). This was evident when comparing the source-target alignment data in the embedded space. Figure 10, bottom, illustrates another instance when this occurs, this time a same-domain alignment scenario.

We also experimented with dynamic mode decomposition (DMD) as an alternative ROM method to infer the target subspace. DMD is known to be superior at capturing the governing temporal structure underlying data. DMD was not suitable for all domain transfer cases mainly since the eigenvectors inferred via DMD are non-orthogonal. It is noteworthy that in the successful cases, we observed significant alignment improvement resulting in longer-term forecasts when compared to POD-derived target subspace. This can be observed by comparing Figure 11 in Appendix B with Figure 12 in Appendix B.

While the demonstrative example in this paper assumes a simple building component, the Physics-based SDA framework is scalable to model more intricate building systems and or thermal zones. There exists dedicated work to automate the generation of RC networks from 3D models of thermal zones (Kim et al., n.d.). Furthermore, the energy balance equations assumed to inform the SSM are well known by the energy community and straightforward to implement.

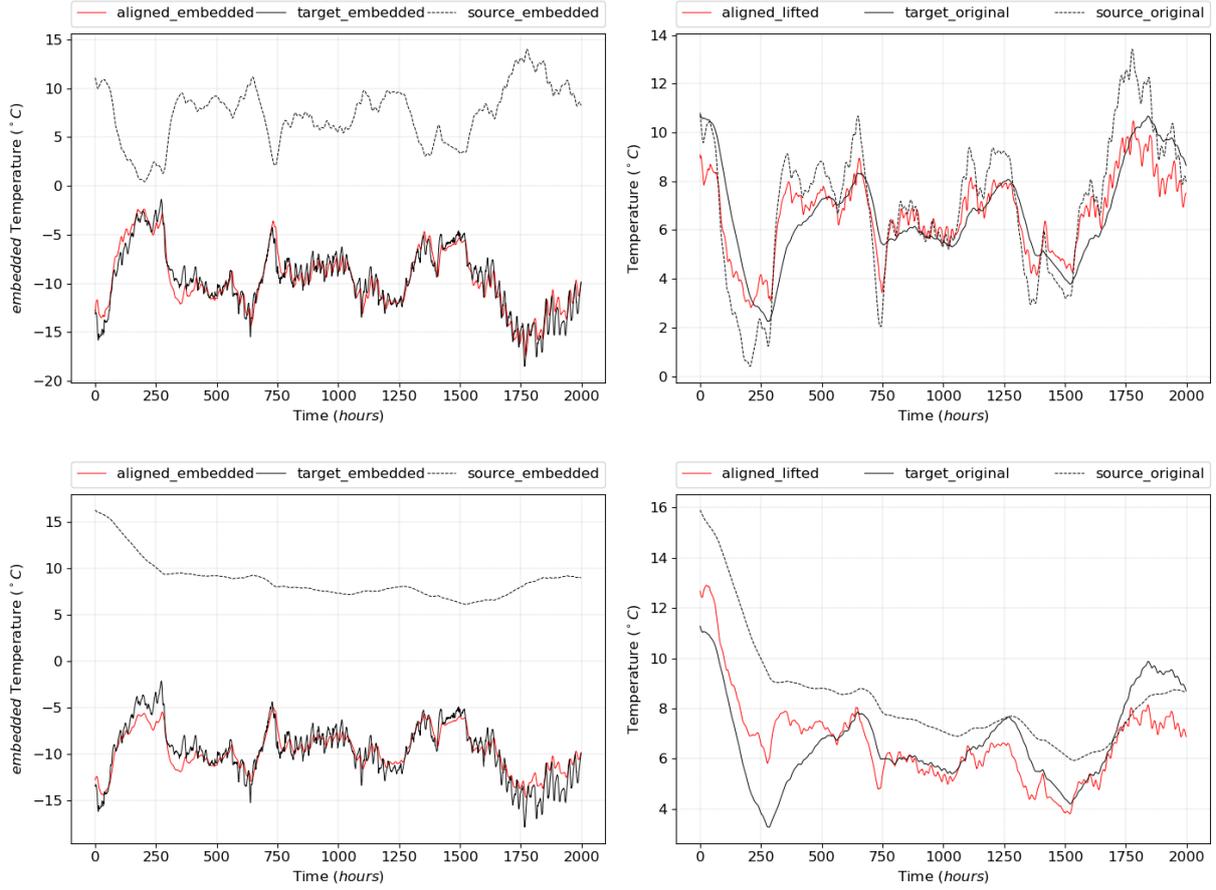

Figure 10. Top: cross-domain alignment: 0.2m wall SSM (source) to 0.9m wall data (target). Middle: same-domain alignment: 0.8m wall SSM (source) to 0.8m wall data (target). Left plots illustrate alignment in the embedded space. Right plots illustrate the alignment after lifting from embedded space.

While the examples illustrated above do not assume noisy conditions in the measurement data, in Appendix C we illustrate a scenario where random noise was added to the measurement data. In noisy measurement conditions, the modelling goal is not to forecast a noisy signal but rather, to forecast a denoised signal. Thus, bypassing a noisy measurement signal wholly depends on the capability of the ROM method to decouple the noise from the true signal. In the example illustrated by Figure 13 and Figure 14 in Appendix C, we introduce noise sampled from a random distribution with a mean of 0.5 and a standard deviation of 0.9, to the measurement data from the scenario discussed in section 5.2. We can observe how the SDA approach is able to learn a transformation between the source and a noise-filtered target signal that generalises successfully beyond measured timesteps. This is largely due to the SVD decomposition of the noisy signal when applying POD to the target data where the SVD acts as a noise filter (Epps & Krivitzky, 2019; Shin et al., 1999). Figure 13 and Figure 14 demonstrate the subspace alignment results in the subspace-embedded and the lifted space, respectively. If POD does not suffice for denoising the timeseries data, there exist extensions of POD such as SPOD (Sieber, M., Paschereit, C. O., & Oberleithner, K., 2016) that focus on spectral decomposition to deal with such decoupling/filtering more successfully. The main difference between POD and SPOD is that the modes vary in both space and time and are orthogonal under space-time inner product, rather than only spatially.

For stochastic LTI SSMs, the presented SDA approach could be extended by introducing a stochastic noise term to a continuous-time LTI SSM which can subsequently be translated in to discrete-time via integration (Madsen, 1995). Such an implementation would not alter the state transition matrix ($A$ matrix) from which we derive the source subspace eigenvectors.

The 'adaptability' of a Physics-based model to forecast for a target measured system using our presented SDA approach is limited by two factors: 1) The state dimensions in the Physics-based model need to be equivalent to the dimensions of the measurement data implying that the resolution of the SSM is limited by the number of available corresponding measurements. In future work we intend to facilitate prediction of unobserved target measurements by leveraging their corresponding Physics-derived state in the SSM; 2) We assume source and target data have pairwise correspondence, implying that the SSM control inputs are the same (measured) control signals known to be influencing the target measurement data, e.g., the external ambient temperature from the case study in section 5. Pairwise correspondence can easily be ensured for MPC applications because most typical building energy scenarios can be described by well-established governing laws, even if approximate.

## 6. Conclusion

In this paper we introduced a novel TL approach to address better model-generalisation for time-series forecasting of building energy, by exploiting well-established ODEs governing thermal transfer, in a data-driven framework. Our approach combines Physics-derived LTI SSMs with subspace alignment where a geometric transformation is sought to align the subspace derived from governing Physics closer towards the subspace decomposed from the measurement data. Subsequently, the transformation is used to forecast beyond the observed measurement data by transforming Physics simulated data for unseen timesteps into approximated measurement data.

While our Physics-based SDA framework addresses generalisation for unseen timesteps (time-series forecasting), it can easily be adopted for cross-system generalisation where for example, we can exploit LTI SSMs identified from data (instead of first principles) of a fully observed system to forecast for unseen timesteps given measurement data from a different but related system. Such an approach could benefit data-efficient control strategies across thermal zones or building components within the same building, by transforming LTI SSM models calibrated on specific thermal zones/components to forecast for other different but related thermal zones/components.

In subsequent work we intend to test the Physics-based SDA framework on real observed scenarios and applications pertaining to energy systems and components in buildings. This work stems from a broader research goal where we aim to overcome persisting caveats with traditional data-driven energy modelling and forecasting, namely generalizability, data-dependency, and interpretability.

Acknowledgments. We are grateful for the technical assistance of Dr Andrew Duncan.

Funding statement. This work was supported by The Alan Turing Institute's Data Centric Engineering Programme under the Lloyd's Register Foundation grant G0095, and Wave 1 of The UKRI Strategic Priorities Fund under the EPSRC Grant EP/T001569/1 and EPSRC Grant EP/W006022/1, particularly the digital twins in engineering theme within those grants & The Alan Turing Institute.
Competing interests. None

Data availability statement. The data used throughout this paper was synthetically generated. The authors have ensured to provide sufficient information to replicate the analytical and numerical models used to generate the synthetic data.
Ethical standards. The research meets all ethical guidelines, including adherence to the legal requirements of the study country.
Author contributions. Conceptualization: Z.X.C; R.C. Formal Analysis: Z.X.C.; L.M. Methodology: Z.X.C. Data visualisation: Z.X.C. Software: Z.X.C. Validation: Z.X.C.; R.C; L.M. Writing original draft: Z.X.C. All authors approved the final submitted draft.

Appendix A: Eigenvectors for each source and target wall type studied in section 5.

Table 5. Eigenvectors for $0.2_{source} \rightarrow 0.2_{target}$

|   | Source (SSM) 0.2m, red brick | | Target (DMD) 0.2m, red brick | | Target (POD) 0.2m, red brick | |
|---|---|---|---|---|---|---|
|   | $V_{s1}$ | $V_{s2}$ | $V_{t1}$ | $V_{t2}$ | $V_{t1}$ | $V_{t2}$ |
| $x$ | 0.985 | 0.172 | -0.837 | 0.011 | -0.687 | -0.727 |
| $y$ | -0.172 | 0.985 | -0.419 | -0.106 | -0.727 | 0.687 |

Table 6. Eigenvectors for $0.6_{source} \rightarrow 0.2_{target}$

|   | Source (SSM) 0.6m, red brick | | Target (DMD) 0.2m, red brick | | Target (POD) 0.2m, red brick | |
|---|---|---|---|---|---|---|
|   | $V_{s1}$ | $V_{s2}$ | $V_{t1}$ | $V_{t2}$ | $V_{t1}$ | $V_{t2}$ |
| $x$ | 0.998 | 0.060 | -0.837 | 0.011 | -0.687 | -0.727 |
| $y$ | 0.060 | 0.998 | -0.419 | -0.106 | -0.727 | 0.687 |

Table 7. Eigenvectors for $0.8_{source} \rightarrow 0.3_{target}$

|   | Source (SSM) 0.8m, red brick | | Target (DMD) 0.3m, concrete | | Target (POD) 0.3m, concrete | |
|---|---|---|---|---|---|---|
|   | $V_{s1}$ | $V_{s2}$ | $V_{t1}$ | $V_{t2}$ | $V_{t1}$ | $V_{t2}$ |
| $x$ | 0.999 | 0.045 | -0.823 | -0.487 | -0.690 | -0.724 |
| $y$ | -0.045 | 0.999 | -0.025 | -0.192 | -0.724 | 0.690 |

Appendix B: Plots for other forecast studies.

Note that for the cross-domain generalisation scenarios in Figure 17 and Figure 12, we assume 3000 hours of training data (vs 2000hours in previous studies) to demonstrate the effect of increased training data on the accuracy of longer-time forecast.

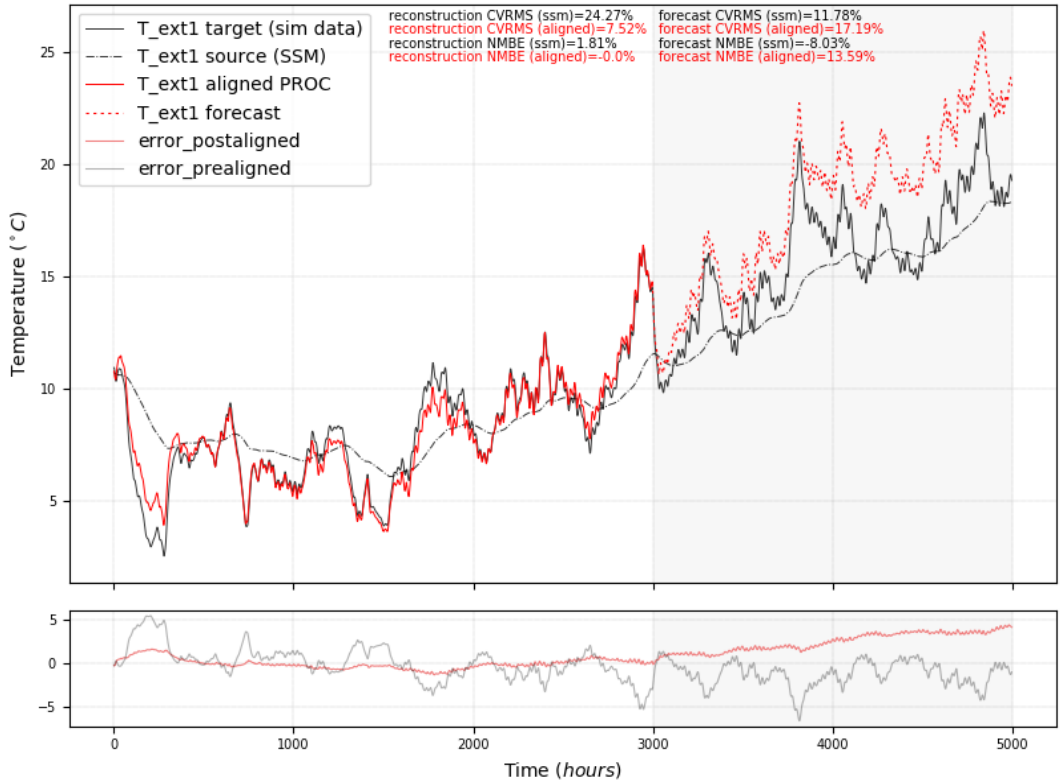

Figure 11. Cross-domain generalization scenario: 0.6m wall SSM (source) to 0.2m wall data (target). The target subspace was derived via POD (orthogonal eigenvectors).

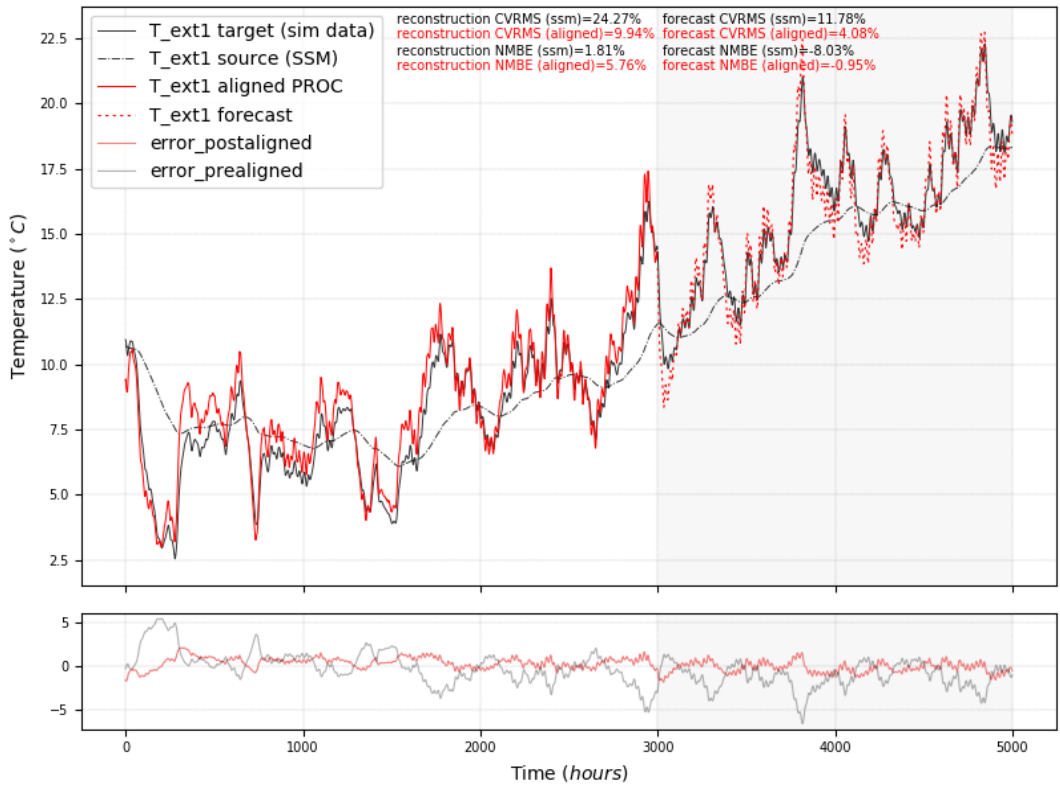

Figure 12. Cross-domain generalization scenario: 0.6m wall SSM (source) to 0.2m wall data (target). The target subspace was derived via DMD (nonorthogonal eigenvectors).

Appendix C: Noisy target measurement data.

We introduce noise sampled from a random distribution with a mean of 0.5 and a standard deviation of 0.9, to the measurement data from the scenario discussed in section 5.2. Figure 13 and Figure 14 demonstrate the subspace alignment results in the subspace-embedded and the lifted space, respectively. We can observe how the Physics-based SDA approach is able to learn a transformation between the source and a noise-filtered target signal that generalises successfully beyond measured timesteps. This is largely due to the SVD decomposition of the noisy signal when applying POD to the target data where the SVD acts as a noise filter (Epps & Krivitzky, 2019; Shin et al., 1999).

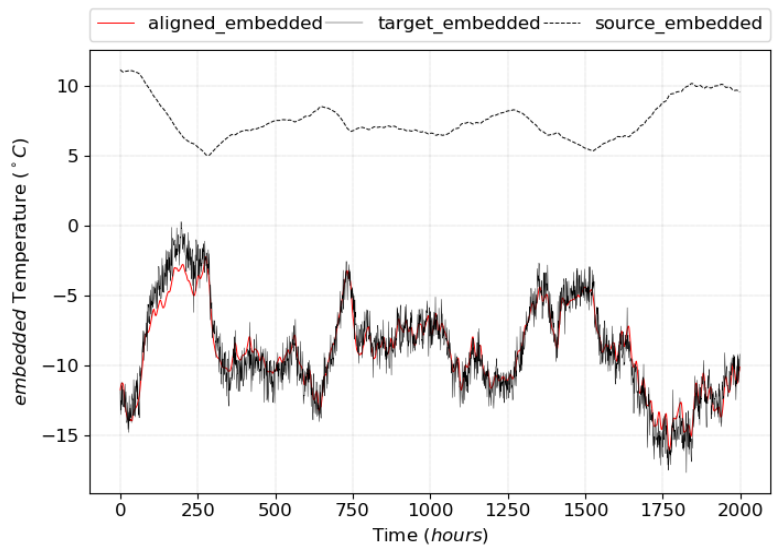

Figure 13. Alignment of source to noisy target signal in embedded space.

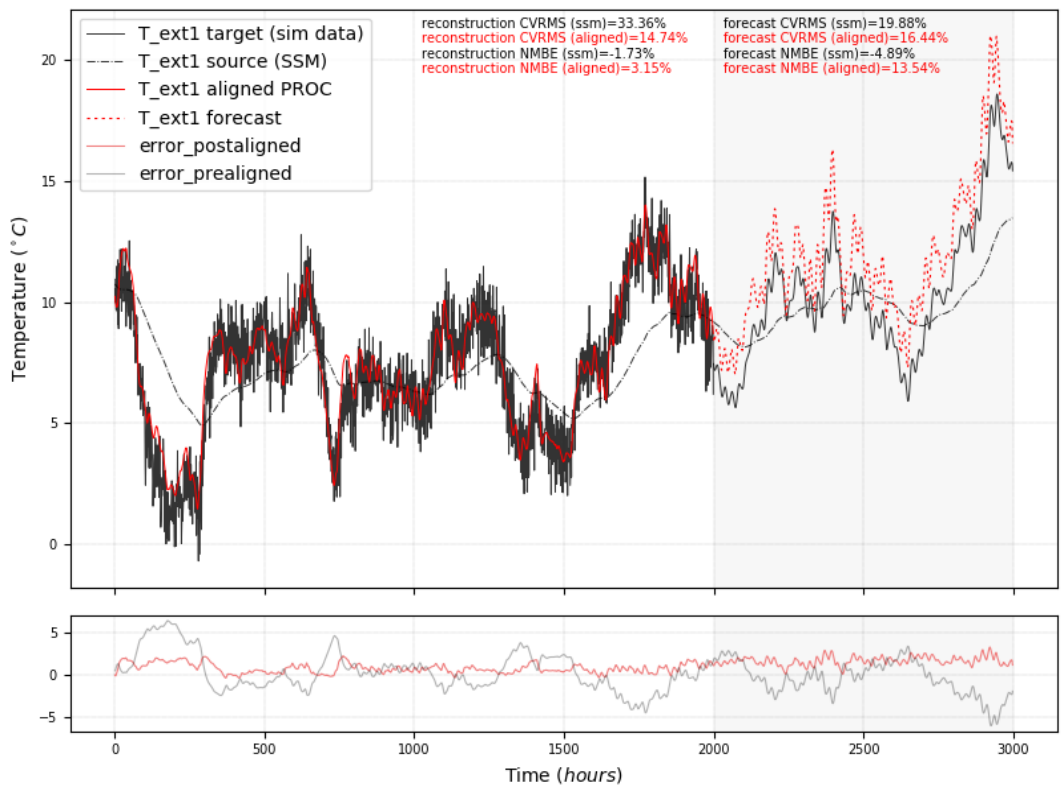

Figure 14. Alignment of source to noisy target signal in lifted space.

Appendix D: Phase portrait plots comparatives between state Physics-derived and data-derived state space geometric structures. Note the geometric similarity between the physics and data.

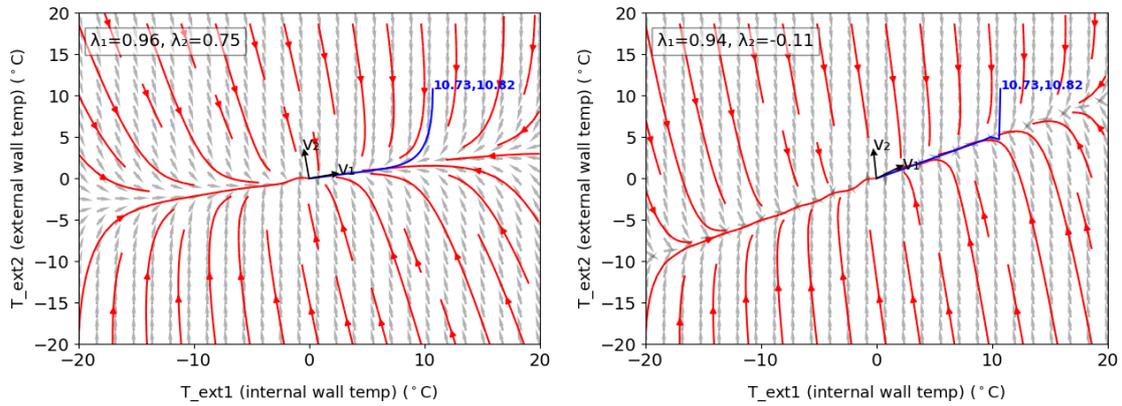

Figure 15. a) Physics-derived state space for 0.2m wall (orthogonal), b) Data-derived state space for 0.2m wall e via DMD (non-orthogonal).

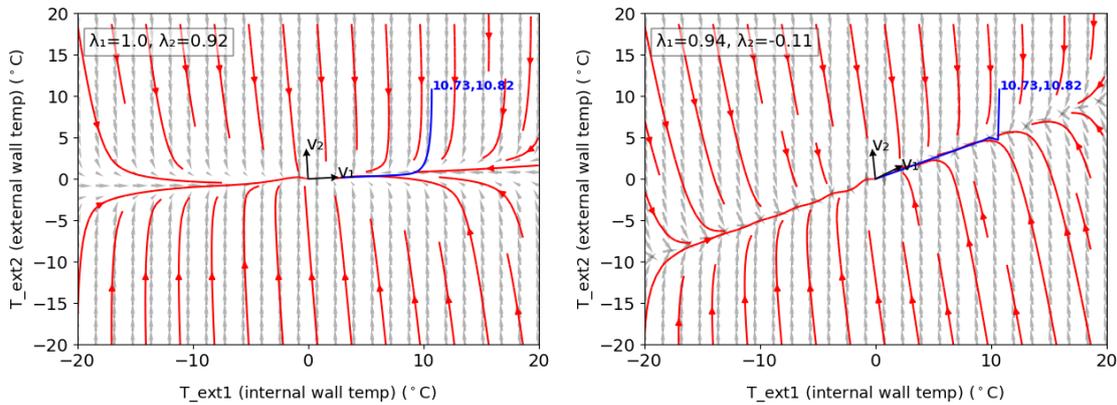

Figure 16. a) Physics-derived state space for 0.6m wall (orthogonal), b) Data-derived state space for 0.2m wall via DMD (non-orthogonal).

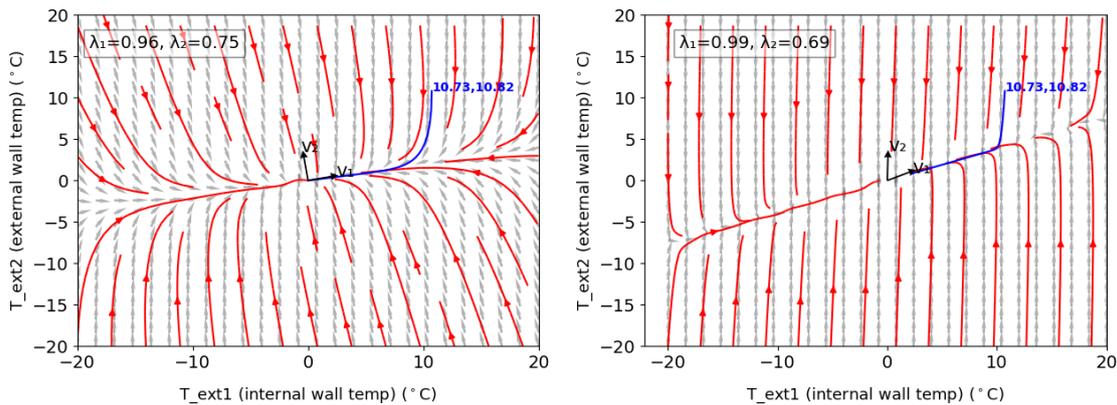

Figure 17. a) Physics-derived state space for 0.2m wall (orthogonal), b) Data-derived state space for 0.9m wall via DMD (non-orthogonal).